\documentclass{article} 
\usepackage{iclr2025_conference,times}

\usepackage{amsmath,amsfonts,bm}

\def\eqref#1{equation~\ref{#1}}

\def\1{\bm{1}}

\DeclareMathAlphabet{\mathsfit}{\encodingdefault}{\sfdefault}{m}{sl}
\SetMathAlphabet{\mathsfit}{bold}{\encodingdefault}{\sfdefault}{bx}{n}

\usepackage{hyperref}
\usepackage{url}

\usepackage{graphicx} 
\usepackage{booktabs}
\usepackage{multirow} 
\usepackage{colortbl}
\usepackage{xcolor}   
\usepackage{threeparttable}
\usepackage{wrapfig}
\usepackage{subcaption}
\usepackage{algorithm}
\usepackage{algpseudocode}
\usepackage{listings}
\usepackage{amsthm}

\usepackage{bbm}
\usepackage{amssymb}
\usepackage{enumitem}
\usepackage{pifont}
\usepackage{tikz}
\usetikzlibrary{shadows}       
\usetikzlibrary{shadows.blur}
\usetikzlibrary{arrows.meta, positioning, shapes, calc, fit}
\usepackage{tabularx}

\definecolor{lightBlue}{RGB}{173, 216, 230}
\definecolor{lightGreen}{RGB}{204, 255, 204}

\definecolor{Blue}{RGB}{0, 183, 133}
\definecolor{Aquamarine}{RGB}{127, 255, 212}
\definecolor{Sepia}{RGB}{112, 66, 20}
\definecolor{BrickRed}{RGB}{203, 65, 84}
\colorlet{my-red}{BrickRed!90!Sepia}
\colorlet{my-blue}{Aquamarine!30!Blue}
\hypersetup{
  colorlinks,
  urlcolor  = blue,
  linkcolor = blue,
  citecolor = blue,
}
\newcommand{\blfootnote}[1]{\begingroup
\renewcommand\thefootnote{}\footnote{#1}\addtocounter{footnote}{-1}
\endgroup}
\usepackage{float}
\usepackage{marvosym}

\title{From Masks to Worlds: A Hitchhiker's Guide to World Models} 

\author{  
  Jinbin Bai$^{1}$,\space\space
  Yu Lei$^{1}$,\space\space
  Hecong Wu$^{1}$,\space\space
  Yuchen Zhu$^{1,2}$,\space\space 
  Shufan Li$^{3}$,\space\space \\
  \textbf{
  Yi Xin$^{1}$,\space\space
  Xiangtai Li$^{1}$,\space\space
  Molei Tao$^{2}$,\space\space
  Aditya Grover$^{3}$,\space\space
  Ming-Hsuan Yang$^{4}$\space\space \vspace{0.15cm}}\\
$^1$MeissonFlow Research $^2$Georgia Tech $^3$UCLA $^4$UC Merced \vspace{0.15cm}\\ 
}

\iclrfinalcopy 
\begin{document}
\maketitle
\blfootnote{\Letter:~ \texttt{jinbin.bai@u.nus.edu}}

\begin{abstract}
This is not a typical survey of world models; it is a guide for those who want to build worlds. We do not aim to catalog every paper that has ever mentioned a ``world model". Instead, we follow one clear road: from early masked models that unified representation learning across modalities, to unified architectures that share a single paradigm, then to interactive generative models that close the action-perception loop, and finally to memory-augmented systems that sustain consistent worlds over time. We bypass loosely related branches to focus on the core: the generative heart, the interactive loop, and the memory system. We show that this is the most promising path towards true world models.
\end{abstract}

\section{Introduction: The Narrow Road to World Models}
The term \emph{world model} has been used to describe many different ideas: learned environment simulators for reinforcement learning \citep{ha2018world, hafner2019dream}, agents that integrate learned models with planning \citep{schrittwieser2020mastering}, and large language models that simulate entire societies \citep{park2023generative}. Yet despite hundreds of related works, there is no clear consensus on how to actually build a true world model. In this paper, we take a stance: the path is much narrower than it appears. 

A true world model is not a monolithic entity but a system synthesized from three core subsystems: a generative heart that produces world states, an interactive loop that closes the action-perception cycle in real time, and a persistent memory system that sustains coherence over long horizons. The history of the field can be understood as an evolutionary journey from first mastering these components in isolation to now integrating them. Most works focus on optimizing narrow tasks and drift away from the generative, interactive, and persistent nature required for a true world model.

\begin{figure}[htbp]
    \centering
    \includegraphics[width=\textwidth]{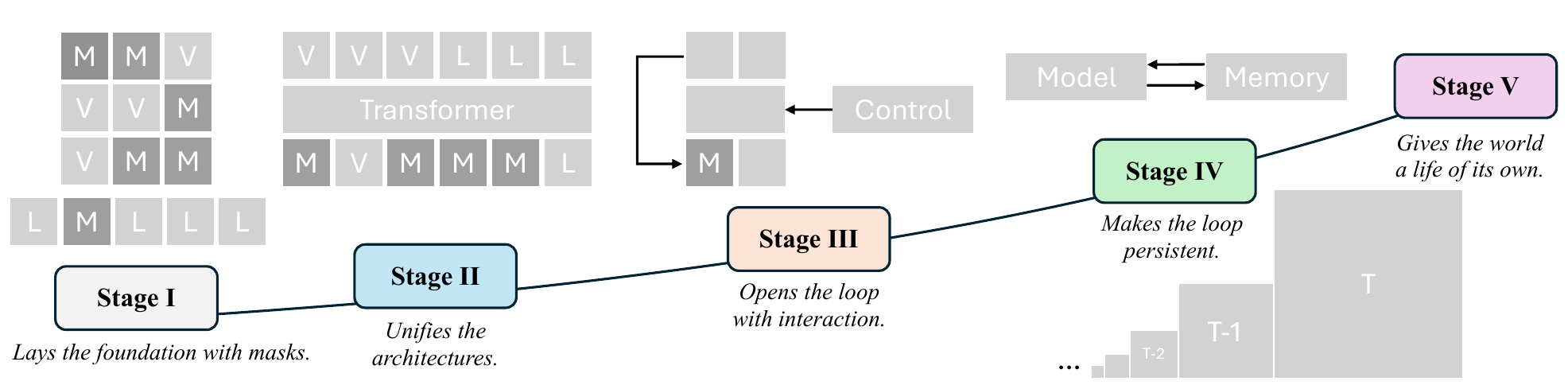}
    \caption{
    The evolution of world models across five stages.
    }
    \label{fig:evolution}
\end{figure}

To make this perspective concrete, we chart the historical evolution of world models as a sequence of five stages, shown in Figure~\ref{fig:evolution}. It begins with Stage I: Mask-based Models, which established a universal, token-based pretraining paradigm across modalities. This foundation enabled Stage II: Unified Models, where a single architecture learns to process and generate multiple modalities. The focus then shifts to closing the interactive loop in Stage III: Interactive Generative Models, transforming static generators into real-time simulators. To sustain these simulations over time, Stage IV: Memory and Consistency introduces mechanisms for durable and coherent state representation. Table~\ref{tab:roadmap_models} also summarizes representative models or methods across the four stages.

\begin{table*}[!ht]
\centering
\caption{Representative models or methods along the narrow road to world models.}
\label{tab:roadmap_models}
\scriptsize
\setlength{\tabcolsep}{4pt}
\renewcommand{\arraystretch}{1.15}
\begin{tabularx}{\textwidth}{>{\raggedright\arraybackslash}p{0.33\textwidth} >{\raggedright\arraybackslash}X}
\toprule
\multicolumn{2}{c}{\textbf{Stage I: Mask-based Models}}\\
\midrule
\textbf{BERT}~\citep{devlin2019bert} & Bidirectional masked prediction for representation learning in language.\\
\textbf{RoBERTa}~\citep{liu2019roberta} & Dynamic masking and scale without next-sentence prediction strengthen BERT.\\
\textbf{Gemini Diffusion}~\citep{deepmind2025gemdiff} & Reported iterative denoising paradigm at commercial scale for generative language tasks.\\
\midrule
\textbf{BEiT}~\citep{bao2021beit} & Image patch masking for representation learning in vision.\\
\textbf{MAE}~\citep{he2022masked} & High-ratio patch masking with lightweight decoder yields strong visual representations.\\
\textbf{MaskGIT}~\citep{chang2022maskgit} & Non-autoregressive parallel masked tokens infilling for efficient image synthesis.\\
\textbf{Meissonic}~\citep{bai2024meissonic} & Masked generative transformers achieving high fidelity text-to-image generation.\\
\midrule
\textbf{wav2vec 2.0}~\citep{baevski2020wav2vec} & Audio latent features masking for representation learning in speech.\\
\addlinespace[0.25em]
\midrule
\multicolumn{2}{c}{\textbf{Stage II: Unified Models}}\\
\midrule
\textbf{EMU3}~\citep{wang2024emu3} & AR-based unified models with a single Transformer for text, image and video.\\
\textbf{Chameleon}~\citep{team2024chameleon} & AR-based unified models with a single Transformer for text and image.\\
\textbf{VILA-U}~\citep{wu2024vilau} & Language-prior AR-based unified models for text, image and video.\\
\textbf{Janus-Pro}~\citep{chen2025janus} & Language-prior AR-based unified models for text and image.\\
\textbf{MMaDA}~\citep{yang2025mmada} & Language-prior mask-based (discrete-style denoising) unified models for text and image.\\
\textbf{Lavida-O}~\citep{li2025lavida} & Language-prior mask-based (discrete-style denoising) unified models for text and image.\\
\textbf{Lumina-DiMOO}~\citep{xin2025lumina} & Language-prior mask-based (discrete-style denoising) unified models for text and image.\\
\midrule
\textbf{UniDiffuser}~\citep{bao2023one} & Visual-prior diffusion-based unified models for text and image.\\
\textbf{Muddit}~\citep{shi2025muddit} & Visual-prior mask-based (discrete-style denoising) unified models for text and image.\\
\midrule
\textbf{UniDisc}~\citep{swerdlow2025unified} & Mask-based (discrete-style denoising) unified models.\\
\midrule
\textbf{Gemini}~\citep{comanici2025gemini25} & Google's multimodal model in a single system (but not in a single paradigm).\\ 
\textbf{GPT-4o}~\citep{hurst2024gpt4o} & OpenAI's multimodal model in a single system (but not in a single paradigm).\\
\addlinespace[0.25em]
\midrule
\multicolumn{2}{c}{\textbf{Stage III: Interactive Generative Models}}\\
\midrule
\textbf{TextWorld}~\citep{cote2018textworld} & Parser-based text game environments.\\
\textbf{AI Dungeon}~\citep{aidungeon2024} & LLM-driven co-authored narrative with open-ended branching stories.\\
\midrule
\textbf{PVG}~\citep{menapace2021playable} & Stepwise playable video game conditioned on user action selection.\\
\textbf{PE}~\citep{menapace2022playable} & 3D playable environments conditioned on camera and multi-object control.\\
\textbf{PGM}~\citep{menapace2024promptable} & Promptable game model conditioned on semantic-level language control.\\
\midrule
\textbf{GameGAN}~\citep{kim2020learning} & GAN-based next frame generation conditioned on actions for 2D games.\\
\textbf{Genie-1}~\citep{bruce2024genie} & MaskGIT-based next frame generation conditioned on actions for 2D worlds.\\
\textbf{Oasis}~\citep{decart2024oasis} & Open-source Diffusion-based real-time generation conditioned on actions for 3D games.\\
\textbf{GameNGen}~\citep{valevski2024diffusion} & Diffusion-based real-time next frame generation conditioned on actions for 3D games.\\
\textbf{Genie-2}~\citep{parker2024genie2} & Diffusion-based generation conditioned on actions for 3D worlds initialized from images.\\

\textbf{Genie-3}~\citep{parker2025genie3} & Real-time generation conditioned on actions and promptable world events for 3D worlds.\\
\textbf{Mineworld}~\citep{guo2025mineworld} & Open-source MaskGIT-based generation conditioned on actions for 3D games.\\
\textbf{Matrix-Game-2}~\citep{he2025matrixgame} & Open-source diffusion-based real-time generation conditioned on actions for 3D games.\\
\midrule
\textbf{World Labs}~\citep{worldlabs2024world} & Explorable 3D environments generation from a single image using geometry and depth.\\
\addlinespace[0.25em]
\midrule
\multicolumn{2}{c}{\textbf{Stage IV: Memory \& Consistency}}\\
\midrule
\textbf{RETRO}~\citep{borgeaud2022improving} & Improving LMs by conditioning on document chunks retrieved from a large corpus.\\
\textbf{MemGPT}~\citep{packer2023memgpt} & OS-inspired virtual memory management framework for LLM workflows.\\
\midrule
\textbf{Transformer-XL}~\citep{dai2019transformer} & Segment-level recurrence with relative positions for long-context sequence modeling.\\
\textbf{Compressive Transformer}~\citep{rae2019compressive} & Extends Transformer-XL by downsampling old states to retain long-range dependencies.\\
\textbf{Mamba}~\citep{gu2023mamba} & Selective state-space model with linear-time recurrence supporting near-infinite context.\\
\midrule
\textbf{FramePack}~\citep{zhang2025packing} & Packs long-frame histories into fixed context with inverted sampling to reduce drift.\\
\textbf{MoC}~\citep{cai2025mixture} & Learnable sparse attention routing that retrieves informative history chunks and anchors.\\
\textbf{VMem}~\citep{li2025vmem} & Introduces surfel-indexed view memory using 3D surfels to enforce spatial coherence.\\

\bottomrule
\end{tabularx}
\end{table*}

This progression culminates in Stage V: True World Models. This stage is not defined by adding a new component, but by the synthesis of the preceding stages into an autonomous whole. At this threshold, models begin to exhibit the defining properties of persistence, agency, and emergence, moving from engines of prediction to living worlds. By analyzing each stage's key innovations and unsolved challenges, this paper offers a clear and opinionated roadmap from today's components to tomorrow's living worlds.

\section{What is a World Model?} \label{sec:definition}

\subsection{Historical and Contemporary Perspectives}
The concept of a world model originated in reinforcement learning, where Ha and Schmidhuber \citep{ha2018world} first proposed learning a latent dynamics simulator for agent planning. This control-oriented view was advanced by systems like Dreamer \citep{hafner2019dream}, which learned policies purely through latent imagination, and MuZero \citep{schrittwieser2020mastering}, which integrated tree-based planning with a learned, abstract model. In parallel, the rise of large-scale generative modeling broadened this definition. With generative agents \citep{park2023generative} and large multimodal systems \citep{reed2022generalist}, the concept evolved from a predictive simulator for an agent to a rich, generative system capable of an entire interactive world. This has led to the contemporary view of a ``world simulator", a term that now informally encompasses three major paradigms: explicit 3D scene generators \citep{worldlabs2024world}, passive video generators that go beyond pixels to approximate physical dynamics \citep{videoworldsimulators2024}, and interactive games and environments for agents, whether text-based \citep{niesz1984interactive} or video-based, as exemplified by the Genie series \citep{bruce2024genie, parker2024genie2, parker2025genie3}.

\subsection{The Anatomy of a True World Model}
\begin{figure}[!ht]
    \centering
    \vspace{-10pt}
    \includegraphics[width=0.8\textwidth]{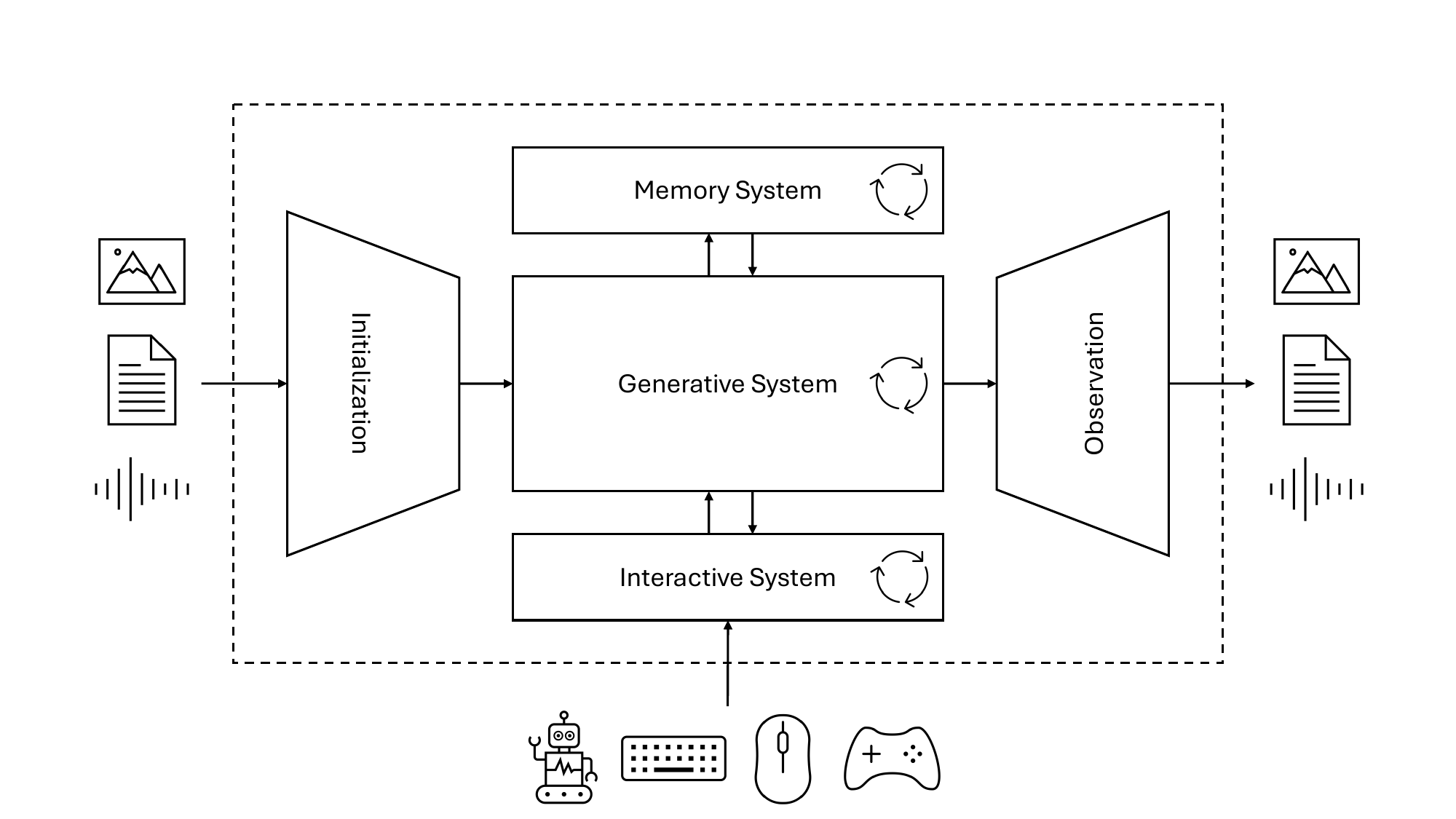}
    \caption{
    The architecture of a true world model. 
    }
    \label{fig:worldmodel-architecture}
\end{figure}

To bring clarity to these diverse threads, we define a true world model as the three essential subsystems it must integrate, which, in turn, enable the core properties that define each stage of our evolutionary roadmap. Figure~\ref{fig:worldmodel-architecture} presents the high-level architecture of a true world model, showing how the generative, interactive, and memory subsystems integrate.

\paragraph{The Generative Heart ($\mathcal{G}$).}
The foundation of a world model is its generative heart: a learned model of the world's dynamics and appearance, formally described by the generative process $p_\theta$. It must be able to predict future states, observations, and the task-relevant outcomes.
\[
\mathcal{G} = \Big(
\underbrace{p_\theta(z_{t+1}\mid z_t,a_t)}_{\text{Dynamics}}, \;
\underbrace{p_\theta(o_t\mid z_t)}_{\text{Observation}}, \;
\underbrace{p_\theta(r_t \mid z_t,a_t)}_{\text{Reward}}, \;
\underbrace{p_\theta(\gamma_t \mid z_t,a_t)}_{\text{Discount/Termination}}
\Big)
\]
This subsystem, which models state transitions, observations, rewards, and terminations, serves as the foundation for the \textbf{Generation} property.

\paragraph{The Interactive Loop ($\mathcal{F}, \mathcal{C}$).}
To be more than a passive movie generator, the model must support a closed interactive loop. For partially observable worlds, it requires an \emph{inference filter} ($q_\phi$) for the agent to interpret observations in real-time, and a \emph{policy} ($\pi_\eta$) for it to act upon its understanding of the world, often paired with a value function ($v_\omega$) to evaluate trajectories.
\[
\mathcal{F}: \quad
\underbrace{q_\phi(z_t\mid h_{t-1},o_t)}_{\text{State Inference}}, \qquad
\mathcal{C} = \Big(
\underbrace{\pi_\eta(a_t \mid z_t,h_t)}_{\text{Policy}}, \;
\underbrace{v_\omega(z_t,h_t)}_{\text{Value}}
\Big)
\]
This loop is what enables true \textbf{Interaction} and \textbf{Real-time Adaptation}.

\paragraph{The Memory System ($\mathcal{M}$).}
Finally, to ensure coherence over time, the model needs a memory system that allows past events to inform the future. This is formally captured by a recurrent state, $h_t$, which is updated based on past memory, the current inferred state, and the last action.
\[
\mathcal{M} : \quad 
\underbrace{h_t = f_\psi(h_{t-1},z_t,a_{t-1})}_{\text{Memory Update}}
\]
This component is the basis for the property of \textbf{Memory}.

\medskip
A detailed formalism of each component is provided in Appendix~\ref{app:definition}. 
This definition clarifies why a system like a Unified Model (Stage II) is a precursor rather than a true world model. While it may possess a powerful generative heart, it typically lacks the dedicated interactive loop and explicit memory system required to sustain a persistent, agent-inhabited world.

\section{Stage I: Mask-based Models Across Modalities}


The first stage in the evolution toward world models is the era of \emph{mask-based modeling}, where a system learns by reconstructing missing or corrupted parts of its input. This paradigm, which can be summarized as mask, infill, and generalize, has proven to be strikingly universal across modalities. It provides a unified way of tokenizing, representing, and pretraining large models, establishing the foundation for all subsequent stages.

\subsection{Language Modality}

Masked language modeling (MLM) has played a foundational role in modern natural language processing. BERT \citep{devlin2019bert} introduced bidirectional context prediction, in which 15\% of tokens in each input are randomly replaced with a [MASK] token and predicted from the surrounding context. SpanBERT \citep{joshi2020spanbert} refined this approach by masking contiguous spans rather than isolated tokens, improving extraction and reasoning tasks. Sequence-to-sequence variants such as MASS \citep{song2019mass}, T5 \citep{raffel2020exploring}, and BART \citep{lewis2019bart} reformulated MLM as a denoising autoencoding objective. ELECTRA \citep{clark2020electra} improved sample efficiency by replacing the MLM objective with a discriminative replacement-detection task.

Beyond fixed-ratio masking, a line of non-autoregressive work introduces dynamic masking and unmasking through iterative refinement. RoBERTa \citep{liu2019roberta} demonstrated that simply optimizing BERT's training recipe with more data and dynamic masking yielded significant gains. Mask-Predict \citep{ghazvininejad2019maskpredict} introduced iterative refinement, re-masking low-confidence tokens over several passes. 
This concept culminated in discrete diffusion models \citep{li2022diffusionlm, he2022diffusionbert, gong2022diffuseq, ou2024your, sahoo2024simple, shi2024simplified}, which replace fixed masking with a time-indexed noise schedule and train the model to iteratively denoise.
As demonstrated by industrial systems like Mercury \citep{inception2025mercury} and Gemini Diffusion \citep{deepmind2025gemdiff}, this dynamic denoising paradigm has matured to rival or exceed autoregressive baselines in both quality and inference speed, solidifying the power of masking as a core generative principle \citep{yu2025discrete,li2025survey}.

\subsection{Vision Modality}

The masked image modeling (MIM) paradigm extended this principle to perception. 
Early works established two main branches. For representation learning, BEiT \citep{bao2021beit} and, especially, MAE \citep{he2022masked} created direct visual analogues of BERT, reconstructing masked tokens or patches to learn powerful features. This spurred a family of related works exploring different reconstruction targets and self-distillation techniques \citep{xie2022simmim, zhou2021ibot, wei2022masked}.

For generative modeling, MaskGIT \citep{chang2022maskgit} and MUSE \citep{chang2023muse} pioneered the use of masked infilling for high-quality parallel image synthesis. This generative trajectory has recently culminated in models like Meissonic \citep{bai2024meissonic}, demonstrating that masked generative transformers can achieve fidelity rivaling that of large diffusion models while offering superior efficiency and control.

This mask-reconstruct-generalize principle scaled effectively to video. VideoMAE \citep{tong2022videomae} and MaskFeat\citep{wei2022masked} showed that high-ratio tube masking was a data-efficient method for learning spatiotemporal representations, confirming that masking could capture not just static scenes but also their dynamics.

\subsection{Other Modalities}
The universality of the masking paradigm was confirmed by its rapid adoption in other fields. In audio, models like wav2vec 2.0 \citep{baevski2020wav2vec}, HuBERT \citep{hsu2021hubert}, WavLM \citep{chen2022wavlm}, and Audio-MAE \citep{huang2022masked} applied masked prediction to latent speech representations. In 3D domains, Point-BERT \citep{yu2022point} and Point-MAE \citep{pang2023masked} adapted masking to point clouds. The principle was even extended to structured data with models like GraphMAE \citep{hou2022graphmae}. These successes reinforced the view of masking as a cross-domain general approach to self-supervised learning.

\medskip
In summary, Stage I established the principle of masking as a universal foundation for representation learning. While this unified the pretraining paradigm, the models themselves remained specialized architectures. The inability of these separate models to form a holistic worldview motivated Stage II: the pursuit of a single, unified architecture.

\section{Stage II: Unified Models} \label{Stage_2}


Stage I established a universal paradigm for representation learning, but the models themselves remained specialists locked within their own modalities. Stage II takes the crucial next step: unifying the models themselves.
We define a unified model as a system that processes and generates across different modalities with a shared backbone and the same paradigm.   
By collapsing modality-specific pipelines, these models simplify scaling, enable powerful cross-modal transfer, and represent the first decisive synthesis on the path toward a true world model.

\subsection{Representative Works}

Leading unified modeling efforts span several trajectories, distinguished by their foundational paradigm. We exclude simple glue models that stitch different paradigms for different modalities, such as using autoregression for text and diffusion for image, as well as models limited to text generation without extending to image generation or other modalities.

\textbf{Extending Language Model Pre-training: Language-Prior Modeling}. 
The dominant trajectory has been to extend the paradigm of autoregressive large language models (LLMs) \citep{radford2019gpt2,brown2020gpt3,sun2024autoregressive}. This began by connecting pre-trained vision encoders to frozen LLMs, as pioneered by BLIP-2 \citep{li2023blip2} and popularized by LLaVA \citep{liu2023llava,liu2024llava15}, which was built upon LLaMA \citep{touvron2023llama}. 
This approach was further pushed into grounded multimodal reasoning by Kosmos-2 \citep{peng2023kosmos} and into embodied reasoning by PaLM-E \citep{driess2023palm}.
More recently, systems like the EMU family \citep{sun2024emu2,wang2024emu3}, Chameleon \citep{team2024chameleon}, VILA-U \citep{wu2024vilau}, and Janus-Pro \citep{chen2025janus} have advanced towards true end-to-end unified generation, creating both text and images within a shared token space and unified autoregressive paradigm.
In parallel, a notable offshoot of this trend is rooted in mask-based language modeling.  
LLaDA \citep{nie2025llada} abandons the autoregressive framework and models text through a masked diffusion process with a single Transformer. Its multimodal extension, MMaDA \citep{yang2025mmada}, introduces a unified discrete diffusion architecture for text and image, a mixed chain-of-thought fine-tuning strategy, and a policy-gradient RL algorithm (UniGRPO) to unify reasoning and generation across modalities within a single model. More recently, Lavida-O \citep{li2025lavida}, OneFlow \citep{nguyen2025oneflow}, and Lumina-DiMOO \citep{xin2025lumina} have further improved overall performance and introduced new capabilities.

\textbf{Extending Vision Model Pre-training: Visual-Prior Modeling}.
A parallel effort, grounded in vision-centric foundations, began along two paths. 
The first path built upon latent diffusion models, the foundation laid by Stable Diffusion \citep{rombach2022high} was later generalized to a unified, joint diffusion process over text and images in models like UniDiffuser \citep{bao2023one} and \citet{rojas2025diffuse}.
The second path is built upon the masked image modeling (MIM) paradigm, with models like Muddit \citep{shi2025muddit} extending Meissonic \citep{bai2024meissonic} into a unified discrete diffusion system that produces both images and captions within a shared architecture and paradigm. Besides, UniDisc \citep{swerdlow2025unified} trained a unified discrete-diffusion model from scratch for both language and vision modalities.

\textbf{Industrial-Scale Unified Systems}. 
At production scale, Gemini \citep{comanici2025gemini25} and GPT-4o \citep{hurst2024gpt4o} unify language and vision modalities in a single model, although not in a single paradigm.  
These demonstrate that unified modeling has transcended research to become a foundational industrial paradigm.

\subsection{Benefits and Gaps}
The primary benefit of Stage II is the reduction of fragmentation, leading to powerful cross-modal transfer and emergent capabilities. 
This paradigm now underpins productized multimodal interaction at scale, as demonstrated by industrial systems like Gemini \citep{comanici2025gemini25} and GPT-4o \citep{hurst2024gpt4o}.
However, despite the impressive progress of language-prior unified models in interactive dialogue, 
visual-prior unified models for text-to-image and text-to-video remain limited to single-shot synthesis or stepwise editing. 
They lack the capacity for continuous, real-time closed-loop interaction. 
Thus, while Stage II unified architectures, the creation of truly dynamic and interactive worlds remains an open challenge and motivates Stage III.

\section{Stage III: Interactive Generative Models} \label{Stage_3}


Here, models are no longer static predictors or one-shot generators, but participants in a closed action-perception loop, sustaining interaction through low-latency response and action-conditioned evolution.  
We define interactive generative models as systems whose outputs are conditioned on streamed inputs or user actions, supported by internal state. 
We explore this evolution across three distinct domains: language-based, video-based, and scene-based.
As mask-based interactive modeling remains underexplored, we take an architecture-agnostic view and summarize general interaction paradigms rather than mask-specific methods.

\subsection{Language-based Worlds: Interaction as Narrative}

Classic interactive fiction (IF) \citep{niesz1984interactive,montfort2011toward,ammanabrolu2020bringing} established the paradigm of text-driven worlds where players interact through textual descriptions and actions. These took several forms:  parser-based games where the player types text commands character by character, choice-based games where the player selects from a set of predefined action options, and hypertext-based games where the player clicks on links embedded in the narrative.
Choice-based visual novels, such as Memories Off \citep{memoriesoff}, exemplify emotionally branching narratives in which player decisions directly affect relationships and endings.
These static worlds naturally evolved into benchmarks for artificial intelligence. A significant line of research, supported by platforms such as TextWorld \citep{cote2018textworld} and Jericho \citep{hausknecht2020interactive}, has focused on training agents to master them. In these settings, the world was a fixed puzzle to be solved, and the locus of intelligence was the agent who navigated a static world, not the world itself.

A fundamental shift occurred when large language models (LLMs) \citep{hurst2024gpt4o,comanici2025gemini25} themselves became the world engine. 
AI Dungeon \citep{aidungeon2024} pioneered this transition, dynamically generating new narrative branches in response to free-form user prompts. Players could explore unbounded story spaces limited only by imagination and the model's generative capacity. 
This marked the transition from solving pre-authored worlds to co-creating open-ended ones, envisioning a future in which visual novels such as Memories Off \citep{memoriesoff} could be generated interactively, offering unique storylines and relationships for each player.

\subsection{Video-based and Scene-Based Worlds: Interaction as Experience} 

Interactive generation in video and spatial domains has progressed from offline frame prediction to real-time, controllable simulation. Early work on world models \citep{ha2018world} used latent rollouts to ``dream" trajectories for policy training, demonstrating the potential of closed-loop simulation. GameGAN \citep{kim2020learning} advanced this idea into a neural game engine that renders successive frames from user input while implicitly learning game rules from observation. 
User control evolved from stepwise action selection in Playable Video Generation (PVG) \citep{menapace2021playable}, 
through 3D scenes with camera and multi-object control in Playable Environments (PE) \citep{menapace2022playable}, 
to natural language prompts in Promptable Game Models (PGM) \citep{menapace2024promptable}, 
which enabled the semantic-level direction of play. 

Building on these conceptual foundations, a decisive trajectory emerged with the Genie series. Genie-1 \citep{bruce2024genie} learned latent action interfaces from Internet-scale videos to create controllable 2D environments. Genie-2 \citep{parker2024genie2} extended this capability to larger, quasi-3D spaces, initialized from a single image and playable via standard controls. Genie-3 \citep{parker2025genie3} scaled further, producing real-time text-to-world experiences at 720p and 24 fps with minutes of coherent play, a marked shift from passive video generation to active interaction. 

Community and industrial efforts soon followed. Systems such as Oasis \citep{decart2024oasis}, GameNGen \citep{valevski2024diffusion}, Mineworld \citep{guo2025mineworld}, and Matrix-Game \citep{he2025matrixgame} demonstrated \textit{real-time} open environments with emergent physics and streaming diffusion. For a comprehensive overview, see the survey by \citet{yu2025survey}.

Beyond frame synthesis, scene-based approaches emerged. World Labs \citep{worldlabs2024world} proposed large world models that generate explorable 3D environments from a single image, enabling interactive navigation through generated geometry and depth rather than sequential video.

Taken together, these advances trace a trajectory from offline video generators to real-time, action-conditioned world simulators. They ultimately transform generative models into engines of interactive human experiences.

\subsection{Challenges}

Despite the leap to real-time interaction, sustaining long-horizon consistency remains unsolved. Two paradigms illustrate the tension: explicit scene generators like NeRFs and Gaussian Splatting (\textit{e.g.}, World Labs) offer stable 3D navigation environments but depend on explicit spatial modeling; implicit frame-by-frame generators offer flexibility but are brittle, prone to losing context and hallucinating objects, especially over extended play. The Genie series highlights this tradeoff: from Genie-1's short 16-frame memory \citep{bruce2024genie}, to Genie-2's object permanence \citep{parker2024genie2}, to Genie-3's few minutes of coherence \citep{parker2025genie3}, progress is clear yet far from persistence. At the object level, implicit video models rely on KV caches or control signals to maintain identity, while explicit 3D approaches embed spatial location directly but still struggle with dynamic elements, as explored in 4D Gaussian Splatting. These challenges reveal a deeper gap: the reactive action–perception loop enables interaction, but without dedicated memory and state management, it cannot sustain persistent worlds — the central theme of Stage IV.

\section{Stage IV: Memory and Consistency}


A world model that acts without memory is reactive yet forgetful.  
This stage aims to endow models with mechanisms that sustain coherent states across long horizons.  
The central question emerges: can world models not only generate but also sustain coherent histories, preserve identities, and resist drift?
We organize this section around three questions: where to anchor memory, how to extend its span and capacity, and how to govern it to preserve consistency. 
Because mask-based persistent memory remains underexplored and varies widely, we take an architecture-agnostic view, focusing on memory and consistency rather than specific mechanisms.

\subsection{Externalized Memory}

Retrieval augments parametric models with non-parametric, often editable, knowledge stores. 
Early explorations such as Neural Turing Machines \citep{graves2014neural}, Differentiable Neural Computers \citep{graves2016hybrid}, and End-to-End Memory Networks (MemN2N) \citep{sukhbaatar2015end} first explored learnable read–write memory slots.
While conceptually groundbreaking, their complexity gave way to more pragmatic, decoupled designs.
kNN-LM \citep{khandelwal2019generalization}, REALM \citep{guu2020retrieval}, and RAG \citep{lewis2020retrieval} showed that conditioning on retrieved passages could dramatically expand effective context while keeping knowledge traceable and updatable.
DPR \citep{karpukhin2020dense} and RETRO \citep{borgeaud2022improving} scaled this approach to dense retrievers and trillion-token databases, rivaling far larger dense LMs while providing traceable and updatable evidence.  

Beyond simple retrieval, research has sought to make memory more scalable and dynamic. 
Product Key Memory (PKM) \citep{lample2019large} supported massive lookup capacity through factorized keys;
MemGPT \citep{packer2023memgpt} reframed LLMs as operating systems with explicit virtual memory management; 
LONGMEM \citep{wang2023augmenting} extends KV caches beyond 65k tokens through decoupled readers;
and From RAG to Memory \citep{gutierrez2025rag} extended retrieval into continual learning, enabling dynamic knowledge updates without retraining.
These systems collectively signal a shift from retrieval as a tool to memory as a co-evolving substrate.

\subsection{Extending Capacity and Span}

Parallel efforts seek to build persistence directly into the architecture, moving beyond fixed-length attention windows. 
Within Transformers, Universal Transformer \citep{dehghani2018universal} introduced depth-wise recurrence; 
Transformer-XL \citep{dai2019transformer} propagated segment states across windows; the Compressive Transformer \citep{rae2019compressive} down-sampled older activations. 
Subsequent designs such as the Memorizing Transformer \citep{wu2022memorizing} and Recurrent Memory Transformer (RMT) \citep{bulatov2022recurrent} 
attached associative key–value stores or persistent memory tokens, reaching million-token horizons in practice; 
Infini-attention \citep{munkhdalai2024leave} added a compressive long-term path for unbounded streaming.
In parallel, Perceiver-AR \citep{hawthorne2022general} introduced a latent cross-attention bottleneck, 
compressing long inputs into a compact representation and enabling autoregression over 100k tokens across text, images, and music.
Together, this line of work represents a reformist trajectory that extends attention through recurrence and compression.

A more radical line argues that persistence requires abandoning quadratic attention entirely. Structured state-space and linear-time models such as S4 \citep{lu2023structured}, Mamba \citep{gu2023mamba}, and RetNet \citep{sun2023retentive} replace attention with recurrent state updates that achieve linear complexity and thereby, in principle, support infinite context.
Precursors such as Linear Transformers \citep{katharopoulos2020transformers}, along with more recent variants such as Hyena \citep{poli2023hyena}, have pointed in this direction by using kernels and long-range convolutions. 
Together, this line of work represents a revolutionary trajectory that abandons attention in favor of continuous dynamical systems.

Scaling strategies and engineering refinements further extend these capacities. 
LongNet \citep{ding2023longnet} employs dilated attention for billion-token contexts; 
Ring Attention \citep{liu2023ring} distributes computation across devices for million-token horizons;
LSSVWM \citep{po2025long} adapts state-space updates for long causal video generation.
Practical techniques such as ALiBi \citep{press2021train}, LongLoRA \citep{chen2023longlora}, and StreamingLLM \citep{zeng2024context} retrofit long-context ability into existing models. 
Together, this line of work represents a pragmatic trajectory that extends persistence through scaling strategies and engineering refinements.

Ultimately, these three trajectories — reformist, revolutionary, and pragmatic — converge on the same goal: achieving genuine continuity, creating models that can read a book, watch a film, or play for hours without losing the thread.

\subsection{Regulating Memory for Consistency}

Persistence without discipline degenerates into drift.  
The nature of this challenge depends critically on the underlying world representation, which has largely followed two paradigms: implicit 2D video frames and explicit 3D scenes.

In implicit, autoregressive video models, the primary challenge is preventing two entangled failures: forgetting, where early content fades, and drifting, where errors compound.  
Efforts to mitigate one often aggravate the other \citep{zhang2025packing}. 
The Genie series highlights this progression:
Genie-1 \citep{bruce2024genie} suffers from short memory and drifts after only a few frames;  
Genie-2 \citep{parker2024genie2} introduces object permanence and sustains coherence for about a minute;
Genie-3 \citep{parker2025genie3} reaches emergent multi-minute consistency.
This underscores a broader challenge: autoregressively generating an environment is fundamentally harder than producing a pre-rendered video, since small inaccuracies accumulate over time.
To tackle this, FramePack \citep{zhang2025packing} uses keyframe anchoring and context compression; Self-Forcing \citep{huang2025self} and CausVid \citep{yin2025slow} impose stronger causal constraints; Context-as-Memory \citep{yu2025context} retrieves overlapping past frames to stabilize long video rollouts, and Mixture of Contexts (MoC) \citep{cai2025mixture} learns sparse routing policies that focus attention on salient history.

Conversely, explicit 3D representations built on generative assets from models like Trellis \citep{xiang2025structured} or TripoSG \citep{li2025triposg} inherently provide strong spatial consistency. Here, the challenge shifts to representing dynamic changes and long-term object states. Methods like WorldMem \citep{xiao2025worldmem}, geometry-grounded spatial memory \citep{wu2025video} and surfel-indexed view memory (VMem) \citep{li2025vmem} leverage this explicit 3D structure to maintain a coherent world state over time, including dynamic representations that capture evolving geometry and supporting revisitations across long horizons. 
Beyond perceptual consistency, maintaining logical and factual coherence in reasoning remains crucial, addressed by techniques that learn to critique their own outputs \citep{asai2024selfrag}.

The overarching lesson is that longer context alone is insufficient. Consistency emerges from explicit policies over memory: what to write, what to retrieve, how to update, and when to forget.

\subsection{Summary}

Stage IV reframes generation as stateful computation.  
Externalized memory makes knowledge editable.  
Architectural persistence makes it durable.  
Consistency policies make it reliable.  
At production scale, multimodal systems such as Gemini \citep{comanici2025gemini25} and Claude \citep{anthropic2024claude35} extend these ideas, sustaining million-token contexts across text, audio, and video and coupling long horizons with reasoning for agentic workflows.  

A deeper question remains.  
Are elaborate memory systems fundamental solutions, or are they sophisticated workarounds for the current constraints of hardware and data?  
The existence of models with massive, brute-force context windows suggests that some memory problems might simply dissolve with sufficient scale, much like how larger models unlocked emergent abilities.
Similarly, consistency failures may also stem from limited data diversity or flaws in the data itself, such as contradictory or erroneous text and videos that are only a few seconds long. 
The answer will determine whether persistence in world models emerges as a natural property of scale, or as the product of carefully engineered memory discipline.
When we ask whether world models can dream consistently, the answer we seek is not just an engineering target but a deeper understanding of the interplay among architecture, scale, and data.

\section{Stage V: Towards True World Models}


The preceding stages constructed the necessary components: a universal generative paradigm (I), a unified architecture (II), a real-time interactive loop (III), and a persistent memory system (IV). Stage V is not the addition of another component, but the synthesis of these parts into a cohesive, autonomous whole. A true world model is not merely a sophisticated simulator controlled by a user; it is a self-sustaining computational ecosystem. Its defining properties are not just programmed but emergent. We show that this synthesis gives rise to three defining properties: Persistence, Agency, and Emergence. 

\subsection{The Threshold: Persistence, Agency, and Emergence}

A true world model ceases to be a program one runs, but a world one enters. Its defining properties are:

\begin{itemize}[leftmargin=*]
    \item \textbf{Persistence}: The world's state and history exist independently of any single user session, accumulating consequences over time. It has a past that can be revisited and a future that unfolds continuously. This is the ultimate fulfillment of the Memory System ($\mathcal{M}$), transforming the property of Memory into an enduring reality.
    \item \textbf{Agency}: The world is inhabited by multiple, goal-directed agents (human or AI) that interact within a shared context. This property is enabled by the Interactive Loop ($\mathcal{F}, \mathcal{C}$), elevating the properties of Interaction and Adaptation into a multi-agent society.
    \item \textbf{Emergence}: The world's macro-level dynamics arise from the micro-level interactions of its agents and underlying rules, rather than being explicitly scripted. The model becomes a crucible for discovering unforeseen social structures, behaviors, and causal chains. This is the critical synthesis that occurs only when the Generative Heart ($\mathcal{G}$), Interactive Loop ($\mathcal{F}, \mathcal{C}$), and Memory System ($\mathcal{M}$) operate in unison over time.
\end{itemize}

\subsection{The Frontier: Three Defining Challenges}

The path to this threshold is defined by three fundamental, unsolved research problems. These are not merely technical hurdles, but grand challenges that constitute the frontier of the field.

\paragraph{The Coherence Problem (Evaluation).}
For conventional models, fidelity is measured against external ground truth. A true world model, however, writes its own history. The challenge is to evaluate the ``truth'' of a self-generating reality: to formalize and measure its internal logical, causal, and narrative coherence, and to define what it means for such a world to be consistent.

\paragraph{The Compression Problem (Scaling).}
An ever-growing history risks computational collapse. The challenge is to learn causally sufficient state abstractions that preserve consequence while discarding noise, approaching the information-theoretic bounds of predictive representation. Yet even with abstraction, long-horizon dynamics may be computationally irreducible, forcing us to treat world models not only as engineered systems but as objects of scientific observation.

\paragraph{The Alignment Problem (Safety).}
An autonomous, persistent world model is a technology with profound societal implications. The alignment challenge for a true world model operates on two distinct levels. At its base, the model itself can be viewed as a single environment whose generating process must align with human values. However, the complexity arises when this model becomes the substrate for a multi-agent society. The alignment problem then becomes squared: it requires aligning not only the world's underlying laws (the substrate), but also the emergent, unpredictable dynamics of the agents interacting within it. This is the harder challenge, distinguishing a true world model from a mere single-environment simulator.

\subsection{The Horizon: From Simulator to Scientific Instrument}

The journey detailed in this paper — from masks to worlds — has been about forging a new kind of technology. Yet, the ultimate promise of a true world model lies beyond its function as a simulator for entertainment or training.

Once a world model crosses the threshold of persistence, agency, and emergence, it transforms from a technological artifact into a new kind of scientific instrument. It becomes a computational crucible for running experiments on complex adaptive systems such as economies, cultures, and cognitive ecosystems that are impossible to conduct in reality.

The quest for true world models, therefore, is not merely an engineering endeavor. It is a pursuit of the ultimate tool for understanding complexity itself. The narrow road leads here: to a future where we build worlds not to escape our own, but to comprehend it.

\section{Conclusion: Building Living Worlds}

This paper has charted a narrow road: a logical progression from the universal paradigm of masking to the threshold of a new reality. We have shown that this path is defined by the sequential mastery of three fundamental capabilities: unified generation, real-time interaction, and persistent memory. These are not ends in themselves, but the necessary foundations for worlds that can truly be called living worlds that persist with their own history, that are inhabited by goal-directed agents, and that give rise to unforeseen emergence.

The pursuit of isolated benchmarks for static tasks is a detour. The true frontier lies in embracing the architectural and theoretical commitments required to build these self-sustaining computational ecosystems. Therefore, the great choice ahead is whether we build worlds as mere tools for entertainment and escapism, or as scientific instruments for comprehending our complexity. The narrow road we have charted leads to this horizon: a future where we forge not just better models, but new mirrors in which to see ourselves.

\clearpage

\appendix

\section*{\LARGE Appendix}
\addcontentsline{toc}{section}{Appendix}
\section{A Formalization of the Three Subsystems}
\label{app:definition}

This appendix provides a detailed breakdown of the components formalized in Section~\ref{sec:definition}. We consider a standard partially observable Markov decision process (POMDP) formulation where at each timestep $t$, an agent takes an action $a_t$, receives an observation $o_t$, and a reward $r_t$. The world terminates based on $\gamma_t$. The model maintains a latent belief state $z_t$ and a deterministic memory state $h_t$.

\paragraph{The Generative Heart ($\mathcal{G}$).} This subsystem models the world's underlying generative process and comprises three components:
\begin{itemize}[leftmargin=*]
    \item \textbf{Dynamics Model} $p_\theta(z_{t+1}\mid z_t,a_t)$: Predicts the next latent state given the current state and an action. This is the core of the model's ability to ``dream" futures.
    \item \textbf{Observation Model} $p_\theta(o_t\mid z_t)$: Maps a latent state back to a sensory observation (\textit{e.g.}, a video frame), grounding the latent space in perceptible reality.
    \item \textbf{Outcome Model} $p_\theta(r_t, \gamma_t \mid z_t,a_t)$: Predicts task-relevant outcomes like rewards and termination signals from the latent state.
\end{itemize}

\paragraph{The Interactive Loop ($\mathcal{F}, \mathcal{C}$).} This subsystem enables a closed-loop exchange between an agent and the world model. It consists of:
\begin{itemize}[leftmargin=*]
    \item \textbf{Inference Model (Filter)} $q_\phi(z_t\mid h_{t-1},o_t)$: Infers the current latent belief state $z_t$ from the new observation $o_t$ and past memory $h_{t-1}$.
    \item \textbf{Control Model (Policy \& Value)} $\pi_\eta(a_t \mid z_t,h_t), v_\omega(z_t,h_t)$: The policy selects the next action based on the current belief and memory, while the value function estimates future outcomes, guiding the policy.
\end{itemize}

\paragraph{The Memory System ($\mathcal{M}$).} This subsystem ensures long-horizon coherence. It has one core component:
\begin{itemize}[leftmargin=*]
    \item \textbf{Memory Update Model} $h_t = f_\psi(h_{t-1},z_t,a_{t-1})$: Updates the deterministic memory state based on the previous memory, the inferred state, and the last action, creating a persistent representation of history.
\end{itemize}

\medskip
This component-based formalization provides a unified lens through which to view the historical evolution of the field, from early control-oriented models that focused on specific components (\textit{e.g.}, \citet{ha2018world}) to modern generative systems that aim to integrate them all. It forms the analytical foundation for the five-stage roadmap presented in this paper.

\clearpage

\bibliography{main}
\bibliographystyle{main}

\end{document}